%% file: main.tex
\PassOptionsToPackage{table}{xcolor}
\PassOptionsToPackage{sort&compress}{natbib}
\documentclass{article}

\usepackage{main,times}
\usepackage[letterpaper,left=0.9in,right=0.9in,top=0.68in,bottom=0.72in]{geometry}
\usepackage{titlesec}

\input{math_commands.tex}

\usepackage[utf8]{inputenc}
\usepackage[T1]{fontenc}
\usepackage{amsmath}
\usepackage{amssymb}
\usepackage{booktabs}
\usepackage{enumitem}
\usepackage{graphicx}
\usepackage{hyperref}
\usepackage{url}
\usepackage{xcolor}

\definecolor{opsdgreen}{HTML}{174B3B}
\definecolor{frontgray}{HTML}{F1F2F3}
\definecolor{frontdark}{HTML}{4B4B4B}
\definecolor{titlestart}{HTML}{3B1C78}
\definecolor{titleend}{HTML}{D9609E}
\newcommand{\gradientmethod}{%
\textcolor{titlestart}{Subje}%
\textcolor{titleend!8!titlestart}{c}%
\textcolor{titleend!20!titlestart}{t}%
\textcolor{titleend!33!titlestart}{A}%
\textcolor{titleend!47!titlestart}{n}%
\textcolor{titleend!61!titlestart}{c}%
\textcolor{titleend!75!titlestart}{h}%
\textcolor{titleend!88!titlestart}{o}%
\textcolor{titleend}{r}%
}
\usepackage{xspace}
\usepackage{tcolorbox}
\usepackage{caption}
\graphicspath{{figures/}{./}}
\setcitestyle{numbers,square,comma,sort&compress}

\hypersetup{
    colorlinks=true,
    linkcolor=blue,
    urlcolor=blue,
    citecolor=blue,
    pdftitle={SubjectAnchor: Subject-Aware Memory-to-Video for Multi-Shot Storytelling},
    pdfauthor={Xinyu Wang, Huafeng Shi, Zian Li, Yan Zhou, Xiaoqiang Liu, Yue Ma, Pengfei Wan},
    pdfsubject={Multi-shot video generation},
    pdfkeywords={video generation, multi-shot storytelling, identity preservation, diffusion transformer}
}

\newcommand{\method}{\texorpdfstring{\textcolor{opsdgreen}{SubjectAnchor}}{SubjectAnchor}\xspace}
\newtcolorbox{frontmatterbox}{
    colback=frontgray,
    colframe=frontgray,
    boxrule=0pt,
    arc=8pt,
    left=20pt,
    right=20pt,
    top=18pt,
    bottom=18pt,
    width=\textwidth,
    before skip=0pt,
    after skip=0pt
}

\titleformat{\section}{\sffamily\Large\bfseries}{\thesection}{1em}{}
\titleformat{\subsection}{\sffamily\large\bfseries}{\thesubsection}{0.8em}{}
\titleformat{\subsubsection}{\sffamily\normalsize\bfseries}{\thesubsubsection}{0.8em}{}
\titlespacing*{\section}{0pt}{2.0ex plus 0.4ex minus 0.2ex}{1.1ex}
\titlespacing*{\subsection}{0pt}{1.7ex plus 0.3ex minus 0.2ex}{0.8ex}
\titlespacing*{\subsubsection}{0pt}{1.4ex plus 0.3ex minus 0.2ex}{0.6ex}

\newcommand{\papersubtitle}{Subject-Aware Memory-to-Video for Multi-Shot Storytelling}
\newcommand{\paperfrontsubtitle}{Subject-Aware Memory-to-Video for Multi-Shot Storytelling}

\title{\method: \papersubtitle}

\newcommand{\frontauthorblock}{%
\textbf{Xinyu Wang\textsuperscript{1,2,$*$},
Huafeng Shi\textsuperscript{2,$\dagger$,$\ddagger$},
Zian Li\textsuperscript{2,3,$*$},
Yan Zhou\textsuperscript{2},
Xiaoqiang Liu\textsuperscript{2},
Yue Ma\textsuperscript{4,$\dagger$},
Pengfei Wan\textsuperscript{2}} \\
\textsuperscript{1}Shenzhen International Graduate School, Tsinghua University \quad
\textsuperscript{2}Kling Team \quad
\textsuperscript{3}Peking University \\
\textsuperscript{4}The Hong Kong University of Science and Technology \\
\texttt{cpwxyxwcp@gmail.com}
}

\newlength{\logoheight}
\newcommand{\logogap}{0.20in}
\newcommand{\affillogo}[2]{%
  \raisebox{-0.5\height}{\includegraphics[height=#1\logoheight]{#2}}%
}

\newcommand{\frontaffiliationlogos}{%
\affillogo{1.00}{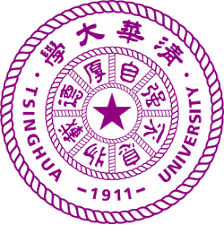}\hspace{\logogap}%
\affillogo{0.92}{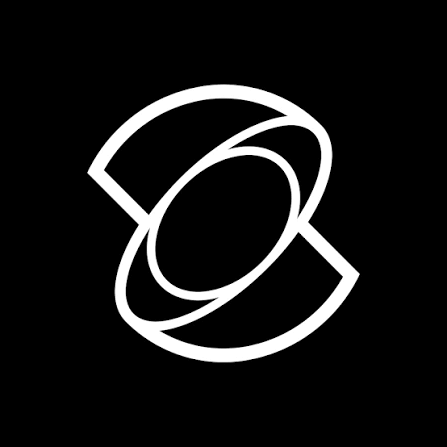}\hspace{\logogap}%
\affillogo{1.00}{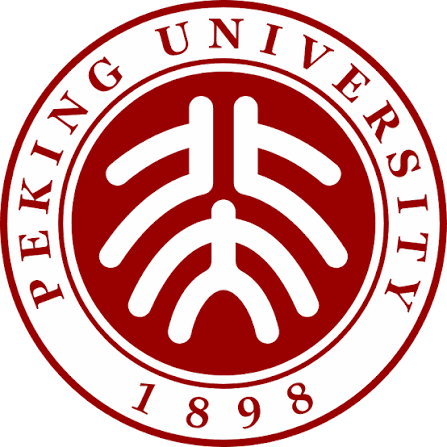}\hspace{\logogap}%
\affillogo{1.10}{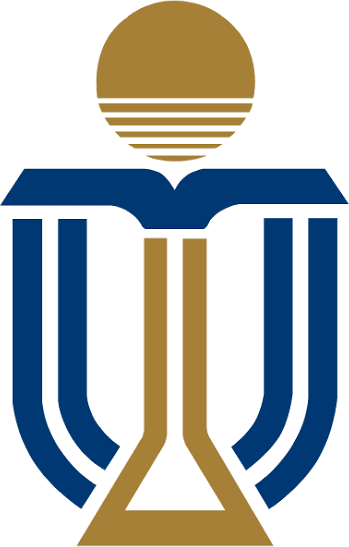}%
}

\newcommand{\makefrontmatter}{%
\vspace*{0.16in}
\begin{frontmatterbox}
{\sffamily\bfseries\fontsize{28}{31}\selectfont \gradientmethod\par}
\vspace{0.08in}
{\sffamily\bfseries\Large\color{frontdark}\paperfrontsubtitle\par}
\vspace{0.06in}
{\sffamily\small\bfseries\textcolor{titlestart}{ACM International Conference on Multimedia (MM '26)}\par}
\vspace{0.18in}
{\normalsize\frontauthorblock\par}
\vspace{0.26in}
{\normalsize\paperabstract\par}
\vspace{0.20in}
\noindent\makebox[\linewidth][r]{\frontaffiliationlogos}
\end{frontmatterbox}
\vspace{0.22in}
}

\arxivcopy 
\renewcommand{\headrulewidth}{0pt}
\renewcommand{\footrulewidth}{0pt}
\AddToShipoutPicture{%
  \AtPageLowerLeft{%
    \raisebox{0.34in}{\makebox[\paperwidth][c]{\thepage}}%
  }%
}

\input{sections/0_abstract}

\begin{document}
\makefrontmatter
{\renewcommand{\thefootnote}{\fnsymbol{footnote}}%
\footnotetext[1]{This work was conducted during the author's internship at Kling AI Research.}%
\footnotetext[2]{Corresponding authors.}%
\footnotetext[3]{Project lead.}%
}

\begin{figure*}[t]
\centering
\includegraphics[width=0.98\linewidth]{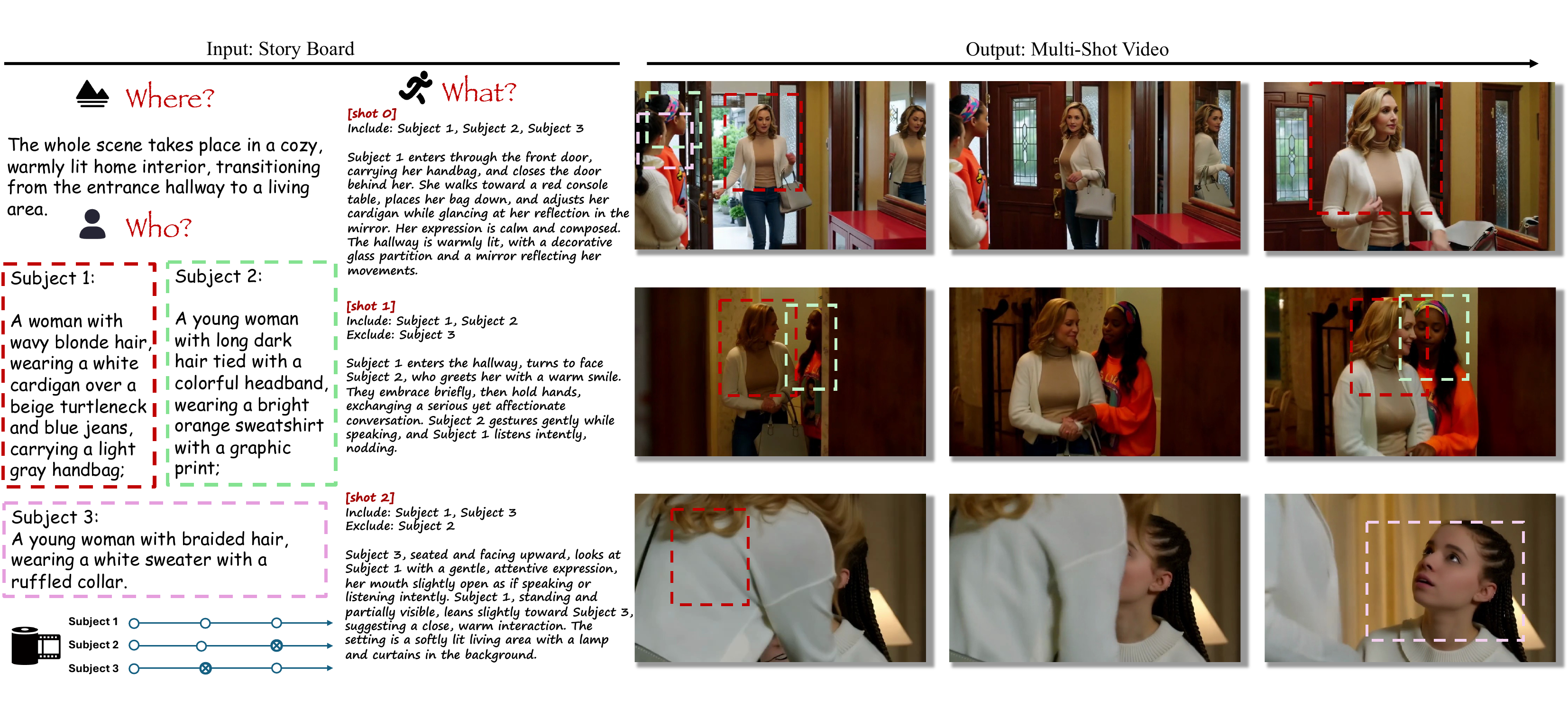}
\caption{\textbf{Showcase of SubjectAnchor.} Our framework decouples complex video narratives into global background/subject settings (``Where'' \& ``Who'') and local shot-by-shot action descriptions (``What''). By explicitly defining subject features and enforcing shot-level inclusion/exclusion constraints, it achieves multi-subject identity consistency and controls complex action interactions across consecutive shots. The dashed bounding boxes in the generated sequences demonstrate the accurate identity mapping of the corresponding subjects.}
\label{fig:teaser}
\end{figure*}

\input{sections/1_introduction.tex}
\input{sections/2_related_works.tex}
\input{sections/3_methodology-v4}
\input{sections/4_experiments-v2}
\input{sections/5_conclusion.tex}

\bibliographystyle{unsrtnat}
\bibliography{main}

\end{document}

%% file: math_commands.tex
\usepackage{amsmath,amsfonts,bm}

\def\eqref#1{equation~\ref{#1}}

\def\1{\bm{1}}

\DeclareMathAlphabet{\mathsfit}{\encodingdefault}{\sfdefault}{m}{sl}
\SetMathAlphabet{\mathsfit}{bold}{\encodingdefault}{\sfdefault}{bx}{n}



%% file: sections/0_abstract.tex
\newcommand{\paperabstract}{%
    We present SubjectAnchor, a Subject-Aware Memory-to-Video paradigm for multi-shot storytelling in which the current shot is generated by conditioning on explicit visual memories extracted from previous shots. The objective is to preserve subject identity and scene consistency across cuts while retaining the controllability of shot-wise prompting. Built on Wan2.2-I2V-A14B, SubjectAnchor contains three key components: subject-related memory construction, subject-aware temporal rotary position encoding, and memory-aware attention partition. For each target shot, the method constructs a compact memory bank by tracing each required subject to its historical appearance and retrieving the most relevant precomputed keyframes. These memory frames are encoded into the model input as explicit visual conditions, while different subjects are assigned to separated negative temporal slots to reduce identity interference. In addition, memory-aware attention partition regulates the interaction between memory tokens and generated content within a shared backbone. This formulation preserves the appearance anchoring of explicit visual memory while remaining compatible with script-driven shot-by-shot generation. Experiments show that SubjectAnchor improves cross-shot identity consistency over representative memory-based and holistic baselines while maintaining competitive visual quality.
}

%% file: sections/1_introduction.tex
\section{Introduction}\label{sec:introduction}
Video generation~\cite{yang2024cogvideox, agarwal2025cosmos, wan2025wan, song2024processpainter, gao2026pai, song2026streamingeffect, song2026vista, kong2024hunyuanvideo, ma2025controllable, li2024survey, an2026vggrpo, an2026video} has advanced rapidly with the emergence of large-scale diffusion transformers~\cite{peebles2023scalable} (DiT), enabling visually realistic clips with rich motion and semantic detail to be synthesized from text, images~\cite{fei2025skyreels, zhu2024champ, ma2026group, he2025magicman, huang2026unityvideo, he2025inpainting}, or reference signals~\cite{jiang2025vace, wang2026customvideo, he2024co, chen2026human, fang20263d}. These advances have greatly improved single-shot video synthesis and brought generative models closer to practical creative workflows. Among the most important downstream applications is cinematic storytelling: real-world films, commercials, and narrative videos are rarely composed of a single continuous shot, but instead unfold through sequences of shots with changing viewpoints, framing, and actions. As a result, \emph{multi-shot storytelling}~\cite{he2025cut2next, meng2025holocine, zhang2025shouldershot} has become a central problem for moving video generation beyond isolated clips toward practical story production.

However, generating a long-form narrative video requires substantially more than producing a set of visually plausible short clips. A practical storytelling system must preserve subject identity, appearance, props, and scene attributes across shot boundaries, while still allowing users to specify each shot through local prompts. This creates a fundamental tension between \emph{global consistency} and \emph{shot-level controllability}. Existing approaches typically resolve this tension in one of two ways.

One line of work adopts \emph{shot-by-shot generation with explicit memory}~\cite{an2025onestory, zhou2024storydiffusion, xiao2025captain, yuan2026helios}. Under this formulation, previously generated shots are converted into visual memory and injected when generating subsequent shots. StoryMem~\cite{zhang2025storymem} is representative of this paradigm. It makes the current shot directly condition on visual evidence from earlier shots rather than inferring identity solely from text. However, such methods still rely on the model to implicitly learn which historical evidence is most relevant for preserving each subject identity. Furthermore, irrelevant memories can easily introduce redundant information, leading to unwanted interference. As a result, memory construction, memory organization, and the interaction between memory tokens and generated tokens remain open challenges, especially in multi-subject scenes.

Another line of work performs \emph{holistic multi-shot generation}~\cite{guo2025long, wu2025cinetrans, wei2025mocha, cai2025mixture, wang2025multishotmaster}, in which all shots in a scene are modeled jointly as a single long sequence. HoloCine~\cite{meng2025holocine} follows this direction. Holistic modeling is attractive for global planning and pacing, but it does not treat explicit memory frames as first-class conditions, which can weaken fine-grained identity anchoring and reduce the modularity needed for local shot editing or regeneration.

Beyond these modeling challenges, progress in this area is also limited by the lack of a benchmark that matches the Memory-to-Video setting. Existing multi-shot video corpora often do not provide the fine-grained structure needed for subject-aware memory retrieval and evaluation, such as story-level persistent descriptions, shot-level subject inclusion and exclusion labels, and subject-centric keyframes that can serve both as memory anchors and as identity-consistency references. Without such annotations, it is difficult to determine whether a model preserves the correct subjects, suppresses irrelevant ones, and follows shot-local instructions while maintaining story-level coherence. More importantly, the absence of a structured benchmark makes it hard to disentangle failures of memory retrieval, identity preservation, and prompt following, which in turn slows down the development of explicit memory-conditioned storytelling systems.

\begin{figure}[t]
\centering
\includegraphics[width=0.98\linewidth]{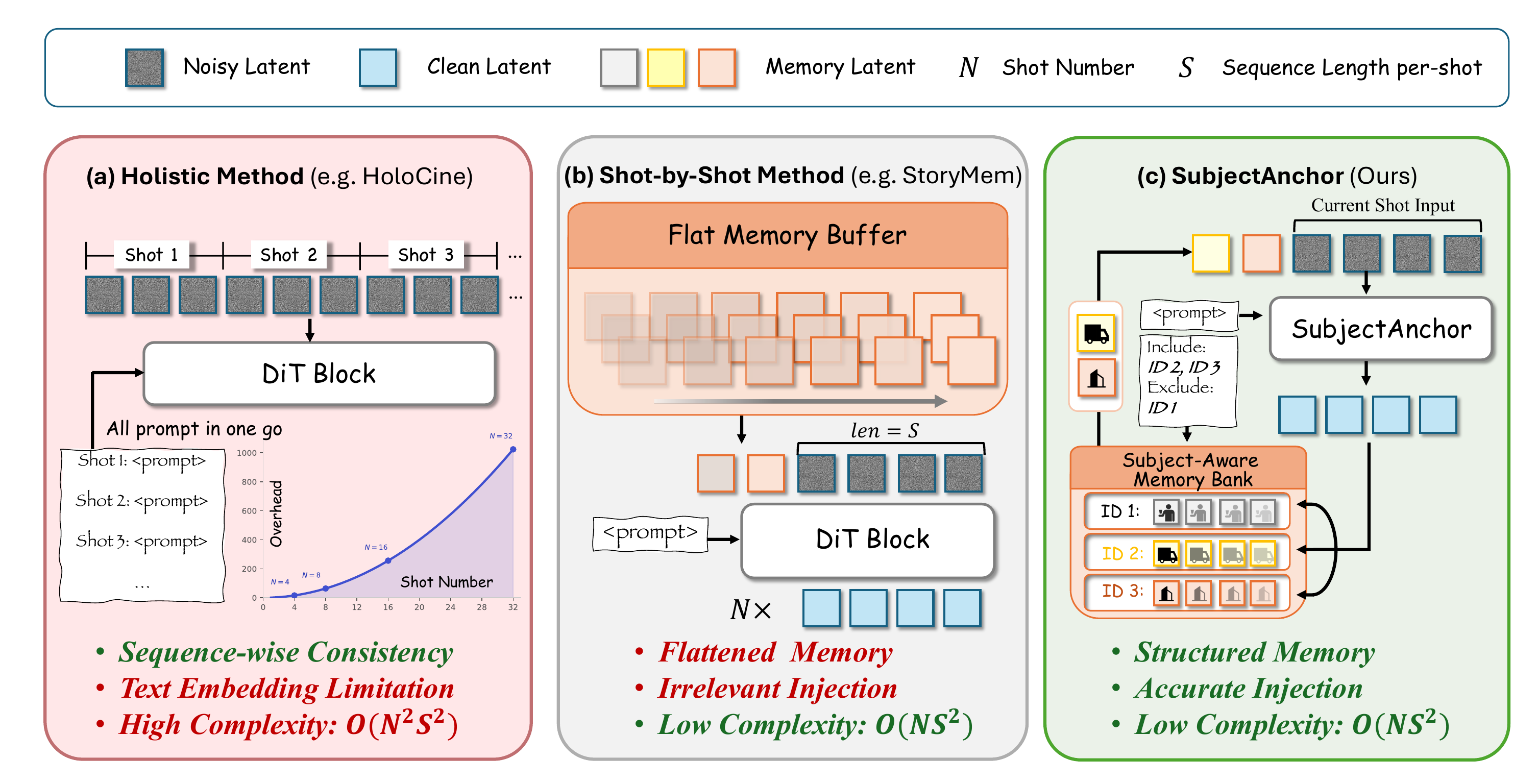}
\caption{\textbf{Comparison of multi-shot video generation paradigms.} (a) Holistic Modeling faces text embedding limitations and an $O(N^2S^2)$ scalability challenge. (b) Previous shot-by-shot methods improve efficiency ($O(NS^2)$) but suffer from a mismatch challenge caused by unstructured FIFO memory buffers. (c) Our SubjectAnchor overcomes these limitations by utilizing a structured, identity-aware memory bank, allowing for selective retrieval of historical frames guided by local inclusion/exclusion constraints.}
\label{fig:architecture}
\end{figure}

To address these challenges, we focus on the first paradigm and study how to improve explicit memory-conditioned generation for multi-shot storytelling. As compared in Figure~\ref{fig:architecture}, unlike holistic approaches that face quadratic computational complexity, and unlike previous shot-by-shot methods that rely on unstructured memory buffers, our approach retains efficient shot-by-shot generation while introducing structured, identity-aware memory control. In contrast to prior memory-based methods that implicitly rely on the model to infer identity-relevant historical evidence, our method makes subject-aware memory selection and organization explicit. Specifically, we build on Wan2.2-I2V-A14B and adapt it into a Subject-Aware Memory-to-Video generator that synthesizes the current shot conditioned on a compact set of visual memories explicitly bound to individual subjects retrieved from previous shots. Ultimately, our framework operates as an autoregressive shot-level generator, naturally scaling to long-form narratives while preserving fine-grained control over individual temporal segments.

This formulation is motivated by three observations. \textbf{i) Not all historical frames are equally useful as memory}: for identity preservation, a compact and subject-targeted memory bank is often more effective than a dense history buffer. \textbf{ii) Memory should be injected explicitly rather than implicitly}: memory frames should be injected as explicit visual conditions rather than treated as loosely correlated context. \textbf{iii) Multi-subject memory requires structured organization}: when multiple subjects coexist in memory, the model benefits from an explicit organization mechanism that separates subject-specific memories in temporal encoding space.

Based on these observations, we propose SubjectAnchor, a Subject-Aware Memory-to-Video framework with three key components: subject-related memory construction, subject-aware memory RoPE~\cite{su2024roformer}, and memory-aware attention partition. Together, these components address which historical memories to retrieve, how to organize multi-subject memories in temporal encoding space, and how memory tokens should interact with generated content inside the shared backbone.

To support large-scale training and systematic evaluation in the setting, we further curate a fine-grained multi-shot storytelling dataset and benchmark with hierarchical annotations, including \textbf{55K} videos for training and a comprehensive evaluation benchmark encompassing diverse narrative scenarios and multi-subject interactions. The dataset provides story-level captions, shot-level subject inclusion/exclusion labels, and subject-centric keyframes, making it possible to jointly assess identity consistency, shot controllability, and subject-aware memory retrieval under recursive multi-shot generation. We then compare SubjectAnchor with representative prior baselines from both memory-based and holistic multi-shot generation paradigms.

Our contributions are summarized as follows:

\begin{itemize}
\item We analyze the memory issue in multi-shot storytelling and propose SubjectAnchor, a practical Subject-Aware Memory-to-Video paradigm with explicit visual memory while preserving shot-level control.

\item We introduce a structured memory mechanism tailored to multi-subject storytelling, including subject-related memory construction for compact identity-preserving retrieval, subject-aware temporal RoPE for organizing multi-subject memories, and memory-aware attention partition for regulating the interaction between memory tokens and generated video tokens within a shared backbone.

\item We design a fine-grained multi-shot storytelling data pipeline, dataset, and benchmark with story-level captions, shot-level subject inclusion/exclusion labels, and subject-centric keyframes, enabling evaluation of identity consistency, shot controllability, and memory retrieval.

\item Extensive experiments demonstrate that SubjectAnchor consistently improves prompt following and cross-shot subject preservation over representative memory-based and holistic multi-shot generation baselines, while remaining competitive in visual quality.

\end{itemize}

%% file: sections/2_related_works.tex
\section{Related Works}\label{sec:related_works}

\subsection{Video Diffusion Transformers}
Recent breakthroughs in video generation~\cite{ma2024followpose,ma2025controllable,ma2024followyouremoji,ma2025followfaster,liu2026opsd,ma2025followyourclick,ma2026fastvmt,ma2025followcreation,ma2025followyourmotion,chen2025contextflow,wan2025unipaint, wang2024taming, yang2025unified, wang2026liveedit, feng2025dit4edit,xu2025clgc,qiu2024tfb, qiu2025duet, qiu2025DBLoss,qiu2026dag, xu2025smrabooth,meng2026parascale} have been largely driven by the transition from U-Net-based architectures to Diffusion Transformers. State-of-the-art models, such as Sora~\cite{brooks2024video}, Cosmos~\cite{agarwal2025cosmos}, and Wan~\cite{wan2025wan}, leverage the scalability of transformers to capture complex spatiotemporal dynamics with high visual fidelity. While these models excel at generating single, continuous clips, they often struggle with long-form narrative consistency. The standard temporal attention mechanism in DiTs is typically confined to a local window or a fixed number of frames, lacking a dedicated interface to anchor long-term visual subjects across multiple scene transitions. Our work extends the Wan2.2-I2V backbone by transforming it into a memory-augmented generator capable of utilizing long historical memory.

\subsection{Identity-Preserving Video Generation}
Preserving subject identity is a fundamental requirement for cinematic storytelling. In the image domain, methods such as IP-Adapter~\cite{ye2023ip}, PhotoMaker~\cite{li2024photomaker}, and InstantID~\cite{wang2024instantid} have established effective pipelines for injecting identity-specific features into diffusion priors. Translating these successes to video, however, introduces the challenge of temporal identity stability~\cite{wu2026language, wu2025dagait, xu2026psgait}. While recent works like ConsistI2V~\cite{ren2024consisti2v} and ID-Animator~\cite{he2024id} attempt to maintain character features within a single shot, they are not designed to handle the abrupt visual shifts and composition changes inherent in multi-shot narratives. SubjectAnchor addresses this by shifting the focus from frame-level identity injection to explicit subject-aware identity control across shots, so that identity preservation is guided by structured memory rather than left to implicitly emerge from historical context.

\subsection{Long Video \& Multi-shot Storytelling}
Narrative video generation can be broadly categorized into holistic modeling~\cite{wu2025cinetrans, wang2025echoshot} and shot-by-shot generation~\cite{an2025onestory, yuan2026helios, zhang2025pretraining} with memory. Holistic approaches, exemplified by HoloCine~\cite{meng2025holocine}, attempt to generate multi-shot sequences as a single coherent latent volume. While effective for global pacing, these methods often lack the modularity needed for local re-editing and can suffer from identity fading in very long sequences. In contrast, memory-based systems such as StoryMem~\cite{zhang2025storymem} generate shots recursively by conditioning each shot on visual evidence from earlier ones. These methods demonstrate the value of explicit memory, but they still rely on the model to implicitly learn which historical evidence~\cite{wu2026promsa} is most relevant for preserving each subject identity. SubjectAnchor follows this second paradigm but introduces explicit subject-aware memory control: memory is selectively constructed, temporally organized, and interactively regulated according to the subject identities required by the current shot. This explicit formulation is particularly important in multi-subject scenes, where identity-specific memory selection and separation remain challenging for existing memory-conditioned frameworks.

%% file: sections/3_methodology-v4.tex
\section{Method}\label{sec:method}
\subsection{Task Definition}
We study \emph{Subject-Aware Memory-to-Video} (SAM2V) for multi-shot storytelling. Given an ordered sequence of story script, the goal is to generate each target shot recursively while preserving subject identity and scene consistency across shot boundaries. For a target shot $S_t$, generation is conditioned on a global caption $g$ that describes persistent story context, a shot caption $c_t$ that specifies the local action and camera behavior, and a subject-related memory set $\mathcal{M}_t$ retrieved from earlier shots.

In the previous memory-conditioned generation methods, historical frames are typically treated as a flat buffer of visual context. In contrast, we explicitly associate memory with the subjects required by the current shot. We denote this required subject set by $\mathcal{U}_t$, and construct $\mathcal{M}_t$ as a subject-related memory bank for the subjects in $\mathcal{U}_t$. This subject-aware formulation is central to our setting: it defines not only what should be generated in the current shot, but also which historical visual evidence should be retrieved to support identity-consistent generation.

Unlike holistic multi-shot generation, we still treat each shot as an independent generation target. Cross-shot consistency is enforced through the shared global caption and the explicit subject-related memory set, which together yield a recursive shot-by-shot storytelling system. In the following, we use $M$ to denote the number of memory frames, $T$ the number of target-video frames, $Z^{\text{mem}}$ and $Z^{\text{vid}}$ the memory and target-video latents, and $X_t$ the joint latent sequence processed by the diffusion transformer.

\subsection{Data Curation}

\begin{figure*}[t]
\centering
\includegraphics[width=1.0\linewidth]{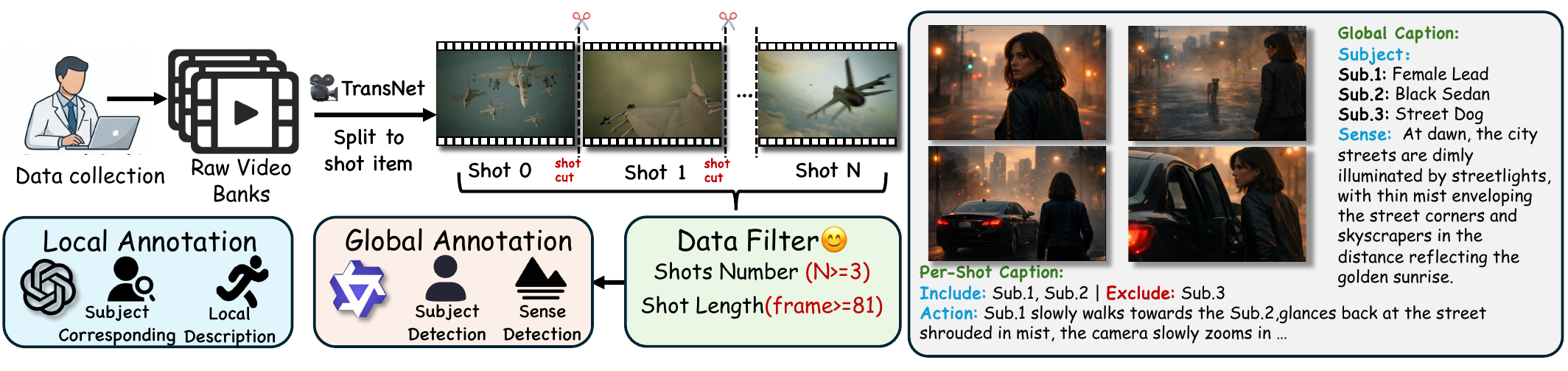}
\caption{Data curation pipeline for Subject-Aware Memory-to-Video. Raw videos are segmented into shots, filtered into valid multi-shot stories, annotated with hierarchical story-level and shot-level descriptions, and paired with subject-centric keyframes. The resulting structured records provide the annotations and memory references used by SubjectAnchor.}
\label{fig:datapipeline}
\end{figure*}

To support Subject-aware Memory-to-Video, we curate the data into structured multi-shot story records rather than treating videos as isolated clips, as shown in Figure~\ref{fig:datapipeline}. Raw videos collected from the Internet and multiple open-source video datasets~\cite{wang2024emu3, wu2025moviebench, song2024moviechat} are first segmented into shots and filtered to retain samples with sufficient multi-shot structure and temporal extent for memory-conditioned generation. In our current setup, we begin with approximately $333$K raw videos, detect shot boundaries with TransNet~\cite{soucek2024transnet}, keep only videos with more than two shots, and further require each shot to contain at least $81$ frames. This yields a curated subset of roughly $55$K videos, in which each target shot can be paired with earlier shots from the same story as candidate memory sources.

Each retained sample is annotated at both the story level and the shot level. We employ Qwen-VL-Max~\cite{bai2025qwen3} to generate these hierarchical annotations from the curated multi-shot videos. At the story level, we maintain a global caption, including a scene description, and a dictionary of subject descriptions. At the shot level, each record contains a shot caption together with explicit include subjects and exclude subjects. Story-level annotations describe persistent scene context and recurring subjects, while shot-level annotations describe local action, framing, and motion. This hierarchical annotation is essential for our formulation because it allows the model to separate persistent subject identity from shot-local behavior.

In addition, we preprocess subject-centric keyframes for the annotated shots to facilitate efficient training-time memory construction. These precomputed keyframes provide compact visual anchors for the subjects appearing in each story, enabling memory retrieval to be organized around subject identity rather than around a flat temporal history. This design not only improves the practicality of large-scale training, but also directly supports the subject-related memory construction, subject-aware RoPE, and memory-aware attention partition used by SubjectAnchor.

\subsection{Overview}
\begin{figure*}[t]
\centering
\includegraphics[width=1.0\linewidth]{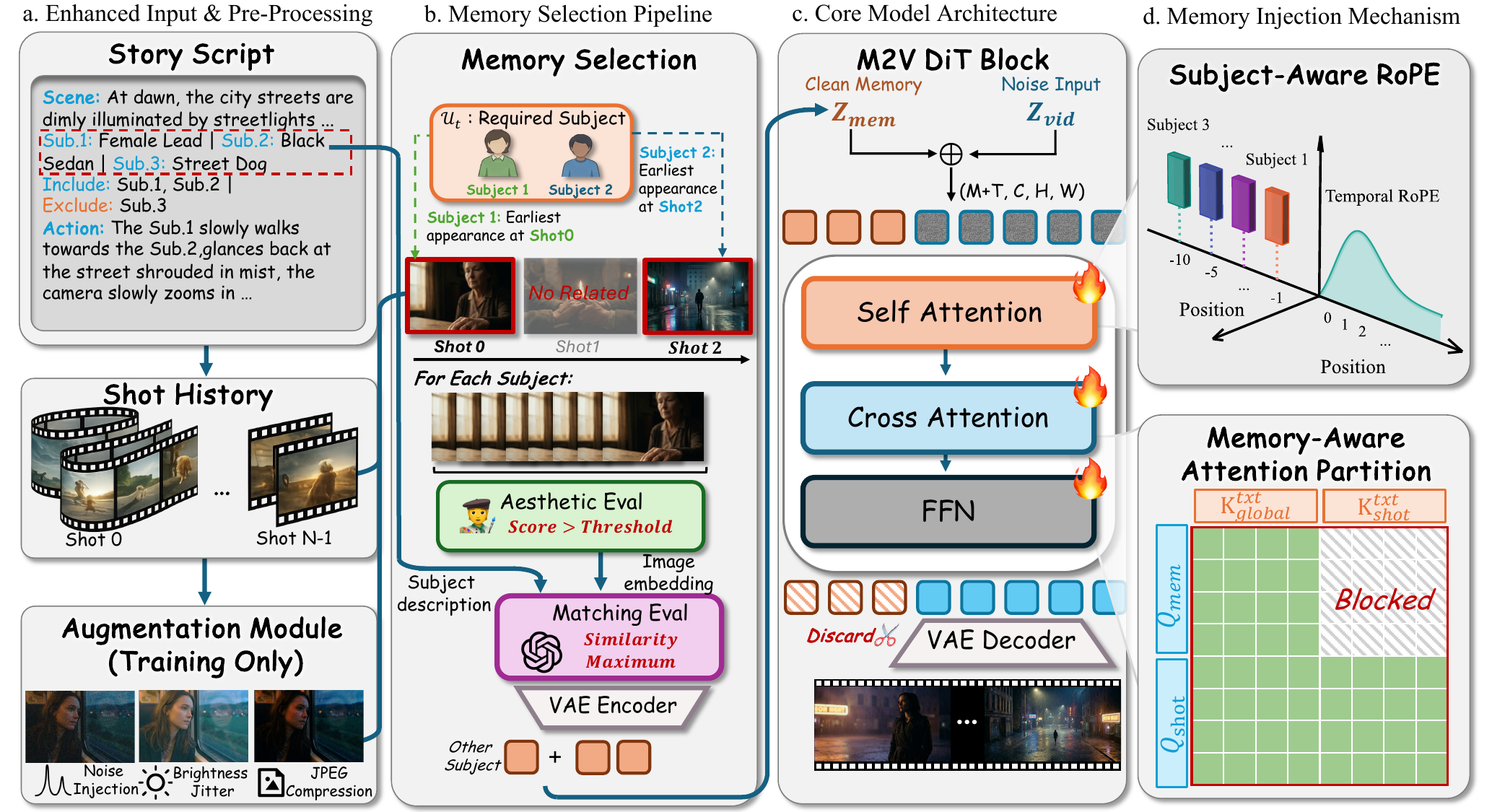}
\caption{Overview of the proposed framework. A target shot first retrieves subject-related memories from previous shots, then encodes them with the VAE, constructs the memory/video mask tensor, injects the resulting representation into the backbone, and organizes memory frames with subject-aware RoPE before diffusion decoding.}
\label{fig:pipeline}
\end{figure*}

Figure~\ref{fig:pipeline} illustrates the overall pipeline. For each target shot, we first construct a compact subject-related memory bank from preceding shots. These memory frames are encoded by the video VAE and concatenated with the target-video latent sequence through a unified memory-conditioned representation. The resulting sequence is fed to the diffusion transformer together with hierarchical text conditions. To better organize multi-subject memories, we assign different subjects to separated negative temporal slots using subject-aware RoPE. The model is then optimized with a flow-matching objective computed only on the target-video segment.

\subsection{Subject-Related Memory Construction}
\paragraph{Earliest-appearance selection.}
For each target shot, we parse the set of required subjects from the shot caption. Our data convention explicitly includes identifiers such as ``\textit{Include: Subject 1, Subject 2}''. Let $\mathcal{U}_t$ denote the required subject set for shot $t$. For each subject $u \in \mathcal{U}_t$, we trace back over previous shots and select the earliest shot in which $u$ appears:
\begin{equation}
\pi_t(u) = \min \{j \mid j < t,\; u \in \mathcal{U}_j\}.
\end{equation}
The final memory source set is the deduplicated collection
\begin{equation}
\Pi_t = \text{unique}\big(\{\pi_t(u) \mid u \in \mathcal{U}_t\}\big).
\end{equation}

This differs from using only the immediately previous shot or simply aggregating all previous shots. The former can miss stable identity references, while the latter introduces redundancy and noise. The earliest appearance serves as a stable anchor because it often provides the cleanest view of a subject before later pose changes, occlusions, or compositional drift accumulate.

\paragraph{Keyframe retrieval.}
Each historical shot in $\Pi_t$ is associated with precomputed subject-centric keyframes and similarity scores. During training, we collect all candidate keyframes from these selected shots, rank them by similarity, and keep the top-$K$ frames. In our setting, the maximum number of memory frames is set to 9. This produces a compact memory set that prioritizes strong identity anchors while keeping computation manageable.

\paragraph{Hierarchical Text Conditioning}
We organize textual conditions into a persistent global component and a shot-local component. Given a global caption $g$ and a shot caption $c_t$, the final prompt is
\begin{equation}
p_t = \textit{``Characters and Scene: } g \textit{''} + \textit{ ``Action: } c_t\textit{''}.
\end{equation}
The global caption describes persistent semantics such as character appearance, wardrobe, and scene layout, while the shot caption describes local motion, action, and camera behavior. This decomposition is important because cross-shot consistency depends not only on local motion prompts but also on persistent identity and scene priors.

\subsection{Memory Injection}
After memory retrieval, the memory frames $\mathcal{M}_t=\{m_1,\dots,m_M\}$ are resized to the target resolution and encoded by the video VAE into memory latents
\begin{equation}
Z^{\text{mem}} \in \mathbb{R}^{C \times M \times H \times W},
\end{equation}
while the target shot is represented by a latent sequence
\begin{equation}
Z^{\text{vid}} \in \mathbb{R}^{C \times T \times H \times W}.
\end{equation}
During training, we further apply memory augmentation to the retrieved frames before VAE encoding, including brightness jitter, additive noise, and JPEG compression, in order to improve robustness to imperfect memory cues.
We inject memory through a unified conditioning design that keeps the original transformer architecture intact while making the role of memory explicit. Specifically, we construct an auxiliary tensor
\begin{equation}
y = \operatorname{concat}_{c}\big(\Omega^{\text{mask}},\; \operatorname{concat}_{t}(Z^{\text{mem}}, Z^{0})\big),
\end{equation}
where $\Omega^{\text{mask}}$ denotes the memory/video mask tensor and $Z^0$ denotes zero placeholders aligned with the target-video segment. Here, $\operatorname{concat}_{t}$ and $\operatorname{concat}_{c}$ denote concatenation along the temporal and channel dimensions, respectively. The tensor $y$ therefore provides the backbone with both the VAE-encoded memory content and an explicit indication of which temporal positions correspond to memory conditions rather than generation targets.

The diffusion transformer then operates on a joint latent sequence
\begin{equation}
X_t = \operatorname{concat}_{t}\big(X^{\text{mem}}_{\text{noise}}, X^{\text{vid}}_{\text{noisy}}\big),
\end{equation}
whose clean counterpart is
\begin{equation}
\bar{X}_t = \operatorname{concat}_{t}\big(Z^{\text{mem}}, Z^{\text{vid}}\big).
\end{equation}
Under this formulation, the memory segment serves as structured conditioning, whereas only the target-video segment is treated as the denoising target. The pair $(X_t, y)$ thus forms the interface between subject-related memory construction and the downstream temporal encoding, attention partition, and masked flow-matching objective. In this way, SubjectAnchor preserves the appearance anchoring provided by retrieved subject-centric frames while retaining a single shared backbone for both memory-conditioned reasoning and target-video generation.

\subsection{Subject-Aware RoPE}
If all memory frames are placed in a single shared negative temporal range, memory tokens from different subjects can become entangled in the rotary position space. This issue becomes more severe in multi-subject stories, where the model must preserve several identities while avoiding interference between their visual memories. To alleviate this problem, we assign each subject a dedicated negative temporal slot and organize the memory bank in a subject-aware manner. Concretely, we assign each subject a slot index $\sigma(u)$ according to its order of first appearance. The $k$-th memory frame of subject $u$ is then assigned temporal position
\begin{equation}
\tau(u,k) = -(\sigma(u)+1)G + k\Delta,
\end{equation}
where $G$ denotes the inter-subject gap and $\Delta$ denotes the intra-subject stride. In our implementation, we set $G=50$ and $\Delta=5$, while target-video frames retain the standard non-negative temporal indices $0,1,\dots,T-1$. This design creates a structured negative-time memory space: memories from different subjects are separated by sufficiently large gaps, whereas multiple frames of the same subject still preserve a local temporal ordering. As a result, the transformer can distinguish subject-specific memory clusters more reliably, reducing identity confusion across subjects while preserving the temporal coherence needed for recursive multi-shot generation.

\subsection{Memory-Aware Attention Partition}
The joint sequence contains two semantically different regions: a memory prefix and a target-video segment. Likewise, the text condition contains a persistent global part and a shot-local part. We preserve this structure by using asymmetric attention rules for memory and target-video queries.

Let $(\mathbf{q}^{\text{mem}}, \mathbf{q}^{\text{vid}})$ denote the memory-query and target-video-query subsets in a transformer block, and let
\begin{equation}
\mathbf{K} = \operatorname{concat}_{t}\big(\mathbf{k}^{\text{mem}}, \mathbf{k}^{\text{vid}}\big), \quad
\mathbf{V} = \operatorname{concat}_{t}\big(\mathbf{v}^{\text{mem}}, \mathbf{v}^{\text{vid}}\big)
\end{equation}
denote the full visual keys and values. Self-attention is defined as
\begin{align}
\mathbf{o}^{\text{mem}} &= \mathrm{Attn}(\mathbf{q}^{\text{mem}}, \mathbf{k}^{\text{mem}}, \mathbf{v}^{\text{mem}}), \\
\mathbf{o}^{\text{vid}} &= \mathrm{Attn}(\mathbf{q}^{\text{vid}}, \mathbf{K}, \mathbf{V}).
\end{align}
Thus, memory queries attend only to the memory prefix, whereas target-video queries attend to the full visual sequence.

For text conditioning, let
\begin{equation}
\mathbf{C} = \operatorname{concat}\big(\mathbf{c}^{\text{global}}, \mathbf{c}^{\text{shot}}\big)
\end{equation}
denote the full text context. Cross-attention is defined as
\begin{align}
\tilde{\mathbf{o}}^{\text{mem}} &= \mathrm{Attn}(\mathbf{q}^{\text{mem}}, \mathbf{c}^{\text{global}}, \mathbf{c}^{\text{global}}), \\
\tilde{\mathbf{o}}^{\text{vid}} &= \mathrm{Attn}(\mathbf{q}^{\text{vid}}, \mathbf{C}, \mathbf{C}).
\end{align}
This design keeps memory tokens as stable conditioning tokens while allowing target-video tokens to use both memory evidence and shot-local text. In implementation, the partition is realized by passing split indices to the shared backbone, which induces grouped attention behavior without introducing separate transformer branches.

\subsection{Training Objective}
We optimize the model with a flow-matching supervised fine-tuning objective on top of the scheduler. At each iteration, we sample a timestep $t$ from the branch-specific noise range and add noise to the target-video segment. The memory prefix is treated as a condition rather than a reconstruction target. Denoting the target-video latent by $Z^{\text{vid}}$ and the sampled noise by $\epsilon$, the noisy target-video input is
\begin{equation}
X^{\text{vid}}_{\text{noisy}} = \mathcal{N}(Z^{\text{vid}}, \epsilon, t),
\end{equation}
and the model predicts the corresponding flow target. The optimization objective is
\begin{equation}
\mathcal{L} = w(t)\cdot \|\hat{v}_\theta - v^\star\|_2^2,
\end{equation}
where $w(t)$ is the scheduler weight, $\hat{v}_\theta$ is the model prediction, and $v^\star$ is the target flow term. Importantly, the loss is computed only on the generated video segment, not on the memory prefix.

Equivalently, the objective can be viewed as a masked flow-matching loss over the joint sequence:
\begin{equation}
\mathcal{L} = w(t)\cdot \left\| \Omega^{\text{vid}} \odot \left(\hat{v}_\theta(X_t, y, p_t) - v^\star\right) \right\|_2^2,
\end{equation}
where $\Omega^{\text{vid}}$ is a binary mask that selects only target-video positions. This makes explicit that memory occupies the same temporal sequence as the generated shot but does not contribute to the reconstruction objective. 

%% file: sections/4_experiments-v2.tex
\section{Experiments}\label{sec:experiments}

\begin{figure*}[ht]
\centering
\includegraphics[width=0.9\linewidth]{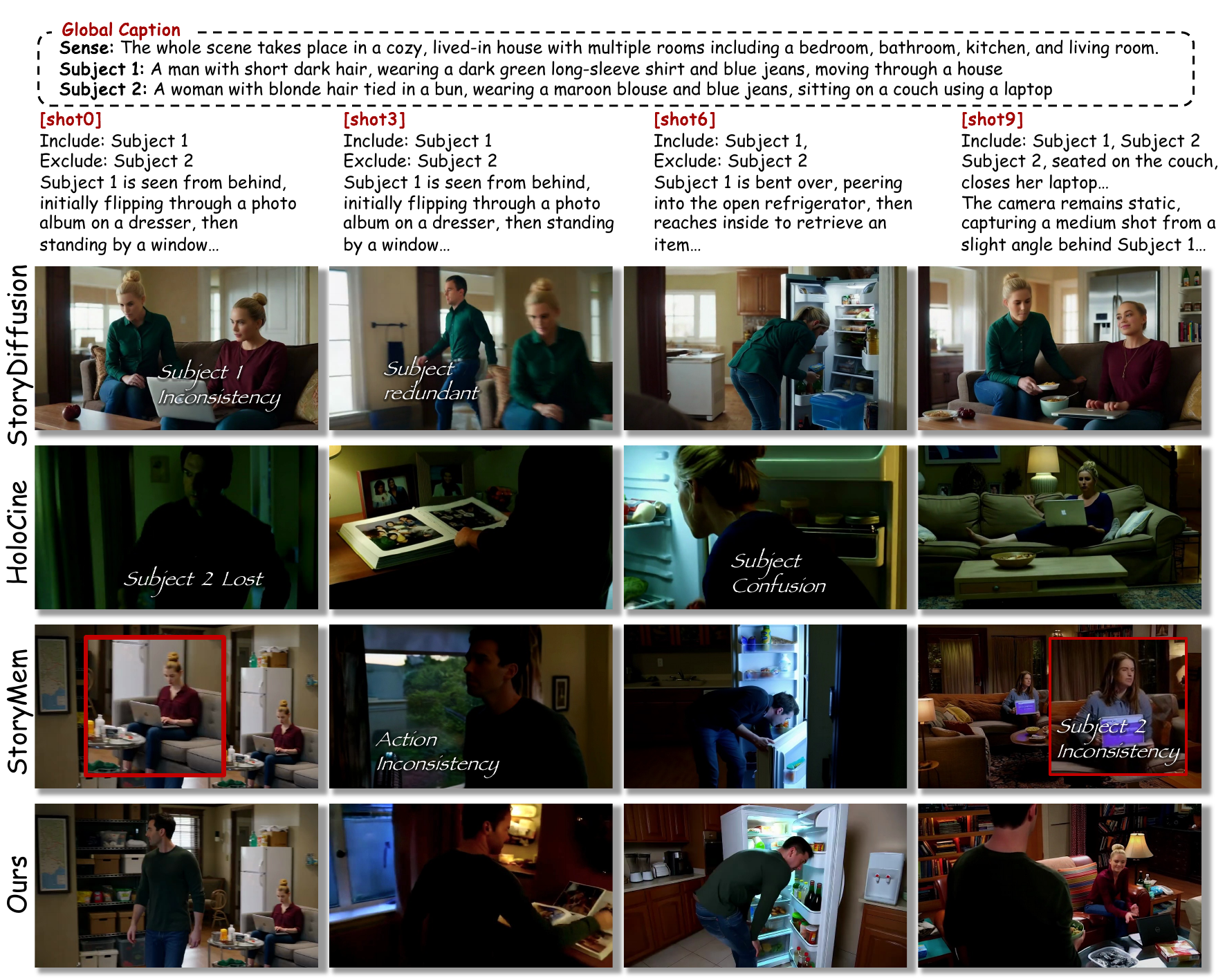}
\caption{Qualitative comparison of multi-shot video generation. SubjectAnchor effectively preserves multi-subject identity and strictly adheres to fine-grained inclusion/exclusion constraints across consecutive shots. Baseline methods struggle with identity inconsistency, subject confusion, and failure to follow local action or exclusion prompts.}
\label{fig:comparison}
\end{figure*}

\subsection{Experimental Setup}
We build SubjectAnchor on Wan2.2-I2V-A14B~\cite{wan2025wan} and adopt a two-stage training strategy for the high-noise and low-noise branches. Both stages use resolution $832 \times 480$, $81$ frames per training sample, and distributed training with a batch size of 32. In both stages, LoRA~\cite{hu2022lora} is applied to Self-Attention, Cross-Attention, and FFN linear layers with rank $128$. The high-noise stage uses timestep range $[0.0, 0.358)$, while the low-noise stage uses the complementary range $[0.358, 1.0]$. 

For memory conditioning, both stages use precomputed subject-centric keyframes retrieved from an offline metadata index, together with global-caption conditioning, memory-aware attention partition, and subject-aware RoPE. The maximum number of memory frames is set to $9$, and memory augmentation is applied with a fixed probability during training. During optimization, only the target-video segment contributes to the flow-matching loss.

\begin{table*}[ht!]
\centering
\caption{Main comparison on multi-shot narrative generation. \textcolor{red}{\textbf{Red}} and \textcolor{blue}{\textbf{Blue}} denote the best and second best results.}
\label{tab:main_results}
\footnotesize
\setlength{\tabcolsep}{3.1pt}
\begin{tabular}{@{}lcccccc@{}}
\toprule
Method & Global Align. $\uparrow$ & Per-shot Align. $\uparrow$ & Subject Align. $\uparrow$ & Identity-First $\uparrow$ & Identity-Prev $\uparrow$ & Aesthetic $\uparrow$ \\
\midrule
StoryDiffusion~\cite{zhou2024storydiffusion}+Wan2.2~\cite{wan2025wan} & 0.3011 & 0.2310 & 0.2353 & \textcolor{blue}{0.8416} & \textcolor{blue}{0.8168} & \textcolor{red}{7.0332} \\
StoryMem~\cite{zhang2025storymem} & \textcolor{blue}{0.3098} & \textcolor{blue}{0.2698} & \textcolor{blue}{0.2548} & 0.7250 & 0.7228 & 6.2093 \\
HoloCine~\cite{meng2025holocine} & 0.2818 & 0.2686 & 0.2284 & 0.4477 & 0.4433 & 4.6530 \\
Ours & \textcolor{red}{0.3243} & \textcolor{red}{0.2729} & \textcolor{red}{0.2814} & \textcolor{red}{0.8689} & \textcolor{red}{0.8681} & \textcolor{blue}{6.5625} \\
\bottomrule
\end{tabular}
\end{table*}

\subsection{Qualitative Comparison}

We compare SubjectAnchor against three baseline paradigms: the two-stage method (StoryDiffusion~\cite{zhou2024storydiffusion} + Wan2.2~\cite{wan2025wan}), the shot-by-shot method (StoryMem~\cite{zhang2025storymem}), and holistic multi-shot generation (HoloCine~\cite{meng2025holocine}). As illustrated in the qualitative comparison (Figure~\ref{fig:comparison}), SubjectAnchor significantly outperforms baseline methods in identity preservation, adherence to exclusion constraints, and action consistency across multi-shot narratives. We observe that two-stage approaches struggle with negative constraints, often leading to subject redundancy, such as the erroneous inclusion of Subject 2 in shot3. HoloCine frequently suffer from ``subject loss'' or character confusion in later frames, failing to distinguish between individuals in complex scenes like shot6. While memory-conditioned baselines improve global coherence, they tend to overfit to historical frames at the expense of local control, resulting in action inconsistencies or severe identity drift by the final shot. In contrast, by leveraging subject-centric memory construction and subject-aware RoPE, SubjectAnchor strictly adheres to inclusion/exclusion prompts and maintains high-fidelity facial features and attire—such as Subject 1's dark green shirt and Subject 2's maroon blouse—even during dynamic transitions like moving from a dresser to a refrigerator. This demonstrates a superior balance between robust visual anchoring and precise local prompt execution. Furthermore, we construct user study, which are provided
in the supplementary materials.

\subsection{Quantitative Comparison}

To complement the qualitative comparison, we construct a subject-structured benchmark for evaluating both prompt alignment and subject consistency. The benchmark contains 50 multi-shot stories and 204 shots in total, with 3-10 shots per story and an average of 4.08 shots. Each story contains 2.84 subjects on average, while each shot involves 1.86 subjects on average. This setup provides a controlled yet challenging evaluation protocol for measuring story-level prompt following and cross-shot identity preservation in multi-subject narrative video generation.

We measure global and local prompt adherence through Global/Per-shot Alignment (CLIP~\cite{radford2021learning} similarity), and assess subject-level fidelity via Subject Alignment. To gauge character constancy across shots, we compute Identity-First and Identity-Prev scores by DINO~\cite{caron2021emerging}, which track feature drifts relative to the first and previous frames. Finally, the LAION-Aesthetic Score Predictor~\cite{schuhmann2022laion} is adopted to provide an objective measure of overall visual quality.

The main quantitative comparison is summarized in Table~\ref{tab:main_results}. As shown in Table~\ref{tab:main_results}, our method achieves the best performance on Global Alignment, Per-shot Alignment, Subject Alignment, DINO Identity-First, and DINO Identity-Prev, indicating stronger prompt following, more accurate subject grounding, and more stable identity preservation across shots. In terms of visual quality, our method obtains the second-best aesthetic score, behind StoryDiffusion+Wan2.2, while maintaining a clear advantage on the identity- and subject-related metrics that are central to multi-shot narrative consistency.

\subsection{Ablation Study}

\begin{figure}[ht]
\centering
\includegraphics[width=0.92\linewidth]{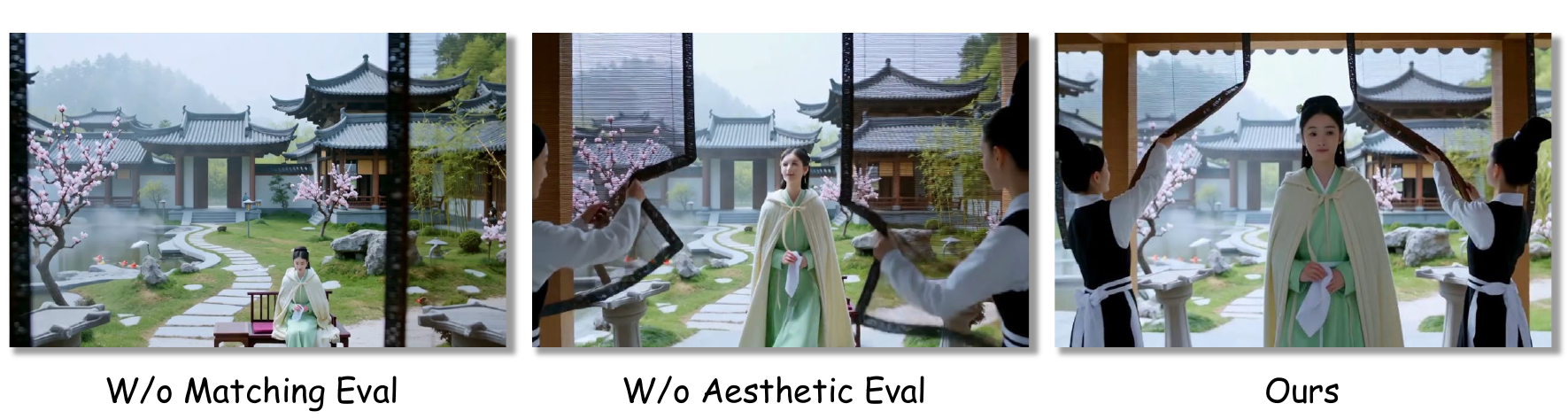}
\caption{Visualization of keyframes under different selection strategies. We compare keyframes retrieved from the same video using three selection strategies. Under ``W/o Matching Eval'', the subject is less salient. Under ``W/o Aesthetic Eval'', the subject is prominent, but the face is blurred. Our method balances subject relevance and frame quality.}
\label{fig:ablation-retrieval}
\end{figure}

\begin{table}[ht!]
\centering
\caption{Memory selection ablation. \textcolor{red}{\textbf{Red}} and \textcolor{blue}{\textbf{Blue}} denote the best and second best results.}
\label{tab:ablation_memory_selection}
\scriptsize
\setlength{\tabcolsep}{3pt}
\begin{tabular}{@{}lcccccc@{}}
\toprule
 & Global Align. $\uparrow$ & Per-shot Align. $\uparrow$ & Subject Align. $\uparrow$ & Identity-First $\uparrow$ & Identity-Prev $\uparrow$ & Aesthetic $\uparrow$ \\
\midrule
W/o Matching Eval & \textcolor{blue}{0.3132} & \textcolor{blue}{0.2514} & 0.2337 & 0.8288 & 0.8276 & \textcolor{blue}{6.4506} \\
W/o Aesthetic Eval & 0.3117 & 0.2461 & \textcolor{blue}{0.2383} & \textcolor{blue}{0.8506} & \textcolor{blue}{0.8513} & 6.4408 \\
Ours & \textcolor{red}{0.3243} & \textcolor{red}{0.2729} & \textcolor{red}{0.2814} & \textcolor{red}{0.8689} & \textcolor{red}{0.8681} & \textcolor{red}{6.5625} \\
\bottomrule
\end{tabular}
\end{table}

\paragraph{Memory selection ablation.}
As shown in Figure~\ref{fig:ablation-retrieval}, ``W/o Matching Eval'' makes it easy to overlook the subject in the selected keyframes, which is difficult to provide enough historical information about the subject for the current shot, leading to a decrease in consistency between shots. Although ``W/o Aesthetic Eval'' can focus on the subject, it is prone to selecting frames with dynamic blurring in the video, which leads to a decrease in memory quality and subsequently lower generation quality. Our method ensures the relevance of characters while also guaranteeing the quality of historical memory, playing a crucial role in maintaining consistency across long temporal sequences for subsequent multi-shot subjects.

\paragraph{Memory injection ablation.}
As shown in Table~\ref{tab:ablation_memory_injection}, removing either subject-aware RoPE or the memory-aware mask leads to a clear degradation in performance, indicating that both components are important for stable memory injection. Without RoPE, the model becomes less effective at organizing subject-specific memory, which weakens identity preservation across shots. Without the memory-aware mask, the distinction between memory and generated video becomes less explicit, resulting in weaker consistency. When both designs are removed, the degradation becomes more pronounced, further confirming that the proposed injection strategy plays an important role in maintaining cross-shot subject consistency. More qualitative comparison results are provided in the supplementary materials.

\begin{table}[ht!]
\centering
\caption{Memory injection ablation. \textcolor{red}{\textbf{Red}} and \textcolor{blue}{\textbf{Blue}} denote the best and second best results.}
\label{tab:ablation_memory_injection}
\scriptsize
\setlength{\tabcolsep}{3pt}
\begin{tabular}{@{}lcccccc@{}}
\toprule
 & Global Align. $\uparrow$ & Per-shot Align. $\uparrow$ & Subject Align. $\uparrow$ & Identity-First $\uparrow$ & Identity-Prev $\uparrow$ & Aesthetic $\uparrow$ \\
\midrule
W/o Subject-Aware RoPE & \textcolor{blue}{0.3196} & 0.2537 & 0.2556 & 0.8476 & 0.8553 & \textcolor{blue}{6.4696} \\
W/o Memory-Aware Attention Partition & 0.3150 & 0.2533 & 0.2507 & \textcolor{blue}{0.8670} & \textcolor{blue}{0.8600} & 6.4181 \\
W/o Both & 0.3165 & \textcolor{blue}{0.2564} & \textcolor{blue}{0.2574} & 0.8097 & 0.8176 & 6.4587 \\
Ours & \textcolor{red}{0.3243} & \textcolor{red}{0.2729} & \textcolor{red}{0.2814} & \textcolor{red}{0.8689} & \textcolor{red}{0.8681} & \textcolor{red}{6.5625} \\
\bottomrule
\end{tabular}
\end{table}

%% file: sections/5_conclusion.tex
\section{Conclusion}\label{sec:conclusion}
We presented SubjectAnchor, a Subject-Aware Memory-to-Video framework for multi-shot storytelling that organizes memory around the subjects required by the current shot rather than treating historical frames as a flat context buffer. Built on Wan2.2-I2V-A14B, SubjectAnchor combines subject-related memory construction, subject-aware rotary position encoding, and memory-aware attention partition to improve identity preservation and shot-level controllability under recursive generation. To support this setting, we also design a fine-grained multi-shot storytelling data pipeline, to curate a structured multi-shot storytelling dataset and benchmark with hierarchical story- and shot-level annotations. Experimental results show that SubjectAnchor improves prompt alignment and cross-shot identity stability while maintaining competitive visual quality, highlighting the value of explicitly structured memory for long-form narrative video generation.